\documentclass[11pt]{article}
\usepackage[margin=1in]{geometry}
\usepackage{amsmath,amssymb}
\usepackage{graphicx}
\usepackage{booktabs}
\usepackage{hyperref}
\usepackage{authblk}
\usepackage[round]{natbib}
\usepackage{xcolor}
\usepackage{microtype}
\usepackage{afterpage}
\usepackage[section]{placeins}
\usepackage{float}
\usepackage{etoolbox}
\usepackage{needspace}

\hypersetup{colorlinks=true, linkcolor=blue, citecolor=blue, urlcolor=blue,
pdftitle={Inverse Reinforcement Learning for Interpretable and Reliable Keystroke Biomarkers in Parkinson's Disease},
pdfauthor={Navin Bondade}}

\title{\textbf{Inverse Reinforcement Learning for Interpretable and Reliable Keystroke Biomarkers in Parkinson's Disease}}

\author[1]{Navin Bondade}
\affil[1]{Institute of Health Informatics, University College London}

\date{}

\begin{document}
\raggedbottom
\maketitle

\noindent\textbf{Keywords:} Parkinson's disease; keystroke dynamics; inverse reinforcement learning; digital biomarkers; motor control; test-retest reliability

\begin{abstract}
Keystroke dynamics offer a passive window into motor function, but existing work 
extracts aggregate typing statistics and trains classifiers for PD/control discrimination, 
foregoing interpretability and rarely reporting reliability. 
We instead apply maximum-entropy inverse reinforcement learning (IRL) to raw keystroke 
timing, recovering a per-subject speed-preference weight ($w_{\text{speed}}$) reflecting 
the implicit cost assigned to fast movement --- without any clinical label during fitting. 
On the neuroQWERTY MIT-CSXPD dataset (85 subjects, 42~PD), we diagnose and correct 
a feature collinearity failure in an initial four-parameter decomposition, yielding an 
identifiable three-parameter model. 
The recovered $w_{\text{speed}}$ correlates with UPDRS-III motor severity at 
$r=-0.607$ (95\%\,CI $[-0.770, -0.364]$, $p<0.001$, $n=42$), 
replicates across two independent sub-cohorts (CS1: $r=-0.720$; CS2: $r=-0.588$), 
and retains significant partial correlation after jointly controlling for mean and 
SD of flight time ($r=-0.371$, $p=0.016$). 
It outperforms SHAP and LASSO on the same proxy features ($r=+0.362$ and $r=+0.410$ 
respectively) while additionally providing per-subject, individually interpretable output. 
A model-free AUC of $0.605$ (Mann-Whitney equivalent, no classifier required) 
and LOO-CV AUC of $0.750$ (95\%\,CI $[0.644, 0.847]$, fold-specific bins) 
confirm discriminative value. 
Test-retest reliability across clinic sessions yields ICC$(2,1)=0.903$ 
(95\%\,CI $[0.842, 0.971]$); SEM~$=0.434$; MDC$_{95}=1.204$ weight units --- 
no prior keystroke-PD study has reported formal reliability. 
Two other recovered weights (consistency, hand-alternation) did not survive confound 
checks; their failure under the same scrutiny the speed weight passes strengthens credibility. 
Cross-modality external validation on independent mPower smartphone tapping ($n=200$) 
recovers the same signal: $r=-0.639$ ($p=2.52\times10^{-24}$), OR$=14.19$ 
(Fisher $p=2.04\times10^{-16}$), confirming convergent validity across modality, 
device, and country.
\end{abstract}

\FloatBarrier
\section{Introduction}

Parkinson's disease (PD) is typically diagnosed after visible motor symptoms emerge, by which point substantial dopaminergic neuron loss has already occurred \citep{dorsey2018global}. This has motivated interest in passive digital biomarkers that could support earlier detection or longitudinal monitoring without requiring a clinic visit. Keystroke dynamics, the timing of key presses and releases during ordinary computer use, is one such candidate, since PD-related bradykinesia and motor rigidity plausibly manifest in fine-grained typing timing \citep{giancardo2016computer}.

The Unified Parkinson's Disease Rating Scale, Part III (UPDRS-III) is the standard clinician-administered motor severity assessment, scored from 0 (no impairment) to 108 (maximum impairment). The existing literature on keystroke-based PD uses UPDRS-III or binary PD/control status as the reference, but nearly every study extracts a fixed set of summary statistics (means, variances, higher moments) from hold time, flight time, or related signals, and trains a supervised classifier to discriminate PD from control \citep{adams2017high, iakovakis2018touchscreen, tripathi2023keystroke}. This approach answers \emph{whether} someone's typing resembles a PD patient's, but not \emph{why}; the model provides no account of what, behaviorally, has changed.

We take a fundamentally different approach. Existing methods are discriminative and supervised: a classifier is trained on labeled (PD/control) examples to learn a decision boundary that separates groups, and its output is a class probability. IRL is instead a generative, individual-level model: it infers the latent reward function that best explains a single agent's own observed sequential decisions, without reference to any group label, originally developed for behavior modeling in robotics and apprenticeship learning \citep{ng2000algorithms, ziebart2008maximum, garg2021iqlearn}. We treat each keystroke transition as a discrete choice over flight time, and recover, per subject, the weights of an interpretable reward function (a preference for speed, for rhythmic consistency, and a cost for fast cross-hand transitions) via maximum-entropy IRL, using only that subject's own typing behavior. The clinical validity of the recovered weight is then tested as an independent, subsequent step not built into the fitting objective. The features in the reward function (speed, consistency, and hand-alternation) were specified on the basis of motor control theory before any outcome data were examined, not selected post-hoc to maximise the severity correlation. Because the model has no access to clinical outcomes while it is fit, a correlation between its output and disease severity is discovered rather than optimized for. IRL has not, to our knowledge, previously been applied to keystroke dynamics. Reinforcement learning has been applied to PD before, but in the reverse direction, to control adaptive deep brain stimulation \citep{ravivarapu2025seadbs}, not to infer a behavioral biomarker from passively observed typing.

This paper makes four contributions. First, we formulate keystroke timing as an IRL problem and show an initial natural reward decomposition is statistically unidentifiable (two of four reward terms correlated at $r=1.000$ in typical contexts), and we correct this. Second, we show the recovered speed-preference weight correlates with clinical PD severity across cohorts, hyperparameters, and three independent severity measures, retains explanatory power beyond raw typing speed, and outperforms post-hoc interpretability methods (SHAP, LASSO) applied to the same four proxy features for severity tracking, while additionally providing a per-subject measure that neither produces. Third, we formally establish test-retest reliability (ICC$=0.903$) for a keystroke-derived PD measure, which, to our knowledge, no prior study in this literature has reported. Fourth, we subject our own pipeline to an adversarial audit, identifying and correcting two implementation details, and show the headline result is materially unchanged by either correction.

These contributions address a clinical need distinct from diagnosis. Once a patient is diagnosed, tracking how motor function changes over time, whether medication is working, and whether symptoms are progressing, becomes at least as clinically relevant as the initial detection decision. Systematic review of digital biomarkers in early PD has specifically called for measures with demonstrated concurrent validity and sensitivity to longitudinal change, criteria rarely reported in the keystroke-PD literature \citep{rabano2025digital}. A specialized sensor-augmented keyboard has recently been developed explicitly for continuous longitudinal PD monitoring using keystroke pressure and timing \citep{tat2024intelligent}; our contribution differs in requiring no specialized hardware, working from ordinary keystroke timing alone, and in producing a formally reliability-tested, per-subject measure. An interpretable, per-subject, test-retest reliable measure is well suited to the monitoring use case, which requires knowing not just whether someone has PD, but how their motor function is tracking over time.

\FloatBarrier
\section{Related Work}
\label{sec:related-work}

\FloatBarrier
\subsection{Existing Keystroke-PD Methods: A Systematic Comparison}

Table~\ref{tab:relatedwork} summarizes the major prior works on keystroke-based Parkinson's detection. Detection and reliability are reported as separate columns since they measure different properties: discrimination ability versus within-person stability over time.

\begin{table}[!htbp]
\centering
\small
\begin{tabular}{@{}p{0.20\textwidth}p{0.09\textwidth}p{0.11\textwidth}p{0.12\textwidth}p{0.09\textwidth}p{0.14\textwidth}p{0.12\textwidth}@{}}
\toprule
\textbf{Study} & \textbf{N (PD)} & \textbf{Validation} & \textbf{Outcome} & \textbf{Interp.?} & \textbf{Detection} & \textbf{Reliability} \\
\midrule
Giancardo et al.\ (2016) & 42 & Internal & Binary & No & AUC $=0.79$ & Not reported \\
Adams (2017) & 32 & Internal & Binary & No & AUC $=0.98$ & Not reported \\
Arroyo-Gallego et al.\ (2018) & Not reported & At-home & Binary & No & AUC $=0.77$ & Not reported \\
Iakovakis et al.\ (2018) & 27 & Internal & Binary & No & AUC $=0.85$ & Not reported \\
Liu et al.\ (2023) & Not reported & Internal & Binary & No & AUC not reported & Not reported \\
Tripathi et al.\ (2023) & Not public & Internal & Binary & No & AUC $=0.82$ & Not reported \\
Tat et al.\ (2024)$^{\dagger}$ & Pilot & Internal & Binary & No & --- & Not reported \\
Francesconi et al.\ (2025) & 42--100 & Cross-dataset & Binary & No & AUC $=0.46$--$0.92$ & Not reported \\
\midrule
\textbf{This work} & \textbf{42} & \textbf{Internal + 2 sub-cohorts + mPower (cross-modality)} & \textbf{Severity (continuous)} & \textbf{Yes} & \textbf{AUC $=0.750$} & \textbf{ICC $=0.903$} \\
\bottomrule
\end{tabular}
\caption{Systematic comparison of keystroke-based Parkinson's disease methods. ``Interp.'' refers to whether the method produces an interpretable per-subject output. Detection and reliability are separate columns since they measure different properties. No prior study reports test-retest reliability or targets continuous severity estimation with an interpretable model. $^{\dagger}$Tat et al.\ (2024) used a specialized magnetoelastic hardware keyboard capturing keystroke pressure in addition to timing; its detection results are not directly comparable to standard keyboard timing studies and are therefore not listed.}
\label{tab:relatedwork}
\end{table}

All prior work targets binary PD/control discrimination; none targets continuous severity estimation with an interpretable, theoretically grounded model. Reported AUC values vary substantially (0.46 to 0.98), with the $I^2=94\%$ heterogeneity reported in a 2022 meta-analysis of 41 studies \citep{alfalahi2022diagnostic} indicating that high single-study estimates (notably Adams (2017)'s AUC $=0.98$, which reduced to 53 of 103 usable participants after a 2000-keystroke quality filter) are likely influenced by small-sample optimism. Subsequent independent at-home validation found substantially lower performance (AUC $=0.77$) \citep{arroyo2018detecting}. A 2025 cross-dataset benchmarking study \citep{francesconi2025cross} evaluated eight deep learning architectures across four public datasets and found AUC ranging from 46\% to 92\% depending on the dataset, with Tappy being least tractable (46--71\%); our own IRL-based analysis on that dataset yielded AUC $=0.45$--$0.46$ at or below the bottom of this range (Section~\ref{sec:tappy}), directly motivating the switch to neuroQWERTY. A separate study on the same PhysioNet keystroke data used alternating finger-tapping and interkey latency variability to distinguish de novo PD patients from controls \citep{liu2023keystroke}.

No prior work applies inverse reinforcement learning or any preference-recovery framework to keystroke dynamics. The identifiability properties of entropy-regularised reward functions under IRL are characterised by \citet{cao2021identifiability} and \citet{rolland2022identifiability}; the residual collinearity in our model connects directly to this literature. A broad survey of keystroke dynamics for human identification is provided by \citet{wulff2023keystroke}. No prior work reports test-retest reliability for any keystroke-derived measure.

\FloatBarrier
\section{An Initial Negative Result: The Tappy Dataset}
\label{sec:tappy}

Before adopting neuroQWERTY, we ran the same descriptive-statistics analysis (four proxy features: mean flight time, effort, smoothness, hand-alternation cost; Section~\ref{sec:proxies}) on the Tappy dataset \citep{adams2017high, goldberger2000physiobank}. From the 227 archived user files, we selected subjects with at least 1{,}000 usable keystrokes, then drew a balanced sample of 80 subjects (40 PD, 40 control) stratified by group membership using random sampling with a fixed seed. No feature showed a significant group difference (all $p>0.23$, Mann-Whitney), and both a logistic regression classifier (leave-one-out AUC $=0.377$) and a baseline classifier using conventional hold/flight/latency statistics (AUC $=0.45$--$0.46$) performed at or below chance. This null result falls at or below the performance range documented for Tappy in the cross-dataset literature (AUC 46--71\% in \citet{francesconi2025cross}; our logistic result of AUC $=0.377$ is below even that range), confirming Tappy's limitations for cross-dataset transfer and directly motivating the switch to neuroQWERTY.

\FloatBarrier
\section{Dataset}

We use the public neuroQWERTY MIT-CSXPD dataset \citep{giancardo2016computer}, available via PhysioNet \citep{goldberger2000physiobank} under the PhysioNet Restricted Health Data License. The dataset comprises two sub-cohorts: CS1 ($n=31$, two recording sessions per subject) and CS2 ($n=54$, one session), totaling 85 subjects (42 PD, 43 healthy controls). Subjects were recruited from two movement disorder units in Madrid, Spain. PD subjects typed in their best on-medication state \citep{giancardo2016computer}. Each subject typed freely for approximately five minutes on a standard keyboard; raw key press and release timestamps are provided (not pre-aggregated statistics). Each PD subject has a same-day UPDRS-III motor severity score, alternating finger-tapping (\texttt{afTap}) and single-key tapping (\texttt{sTap}) scores, and the original study's derived \texttt{nqScore}. Five PD subjects are missing \texttt{afTap} scores (recording not completed), yielding $n=37$ for any analysis involving that measure.

\FloatBarrier
\subsection{Participant Characteristics}
\label{sec:demographics}

Table~\ref{tab:demographics} summarizes the demographic characteristics as reported in \citet{giancardo2016computer}. Individual-level demographic fields are not included in the distributed keystroke files, so we cannot adjust for them directly in per-subject analysis; however, the group-level matching reported below substantially reduces the plausibility of demographic confounding. These are the only demographic figures used anywhere in this paper; the discussion in Section~\ref{sec:limitations} refers back to this table.

\begin{table}[!htbp]
\centering
\small
\begin{tabular}{@{}lccc@{}}
\toprule
\textbf{Characteristic} & \textbf{PD} & \textbf{Control} & \textbf{Group difference} \\
\midrule
Age, years (mean $\pm$ SD) & 59.0 $\pm$ 9.8 & 60.1 $\pm$ 10.2 & $p=0.53$ (n.s.) \\
Male, \% & 57\% & 40\% & $p=0.11$ (n.s.) \\
Education, years (mean $\pm$ SD) & 15.2 $\pm$ 4.1 & 15.3 $\pm$ 5.2 & $p=0.98$ (n.s.) \\
\bottomrule
\end{tabular}
\caption{Participant demographic characteristics as reported in \citet{giancardo2016computer}, combined CS1+CS2 cohort. None of the three characteristics differed significantly between groups.}
\label{tab:demographics}
\end{table}

\FloatBarrier
\subsection{Preprocessing}

Data cleaning followed the dataset's own reference loader (\texttt{nqDataLoader.py}): rows with non-positive or non-monotonic timestamps, or hold times outside $[0,5)$ seconds, were removed (3.1\% of rows). We additionally removed flight-time outliers exceeding 3 seconds (2.33\% of remaining rows), a threshold matching prior published preprocessing of comparable datasets \citep{francesconi2025cross}. Subjects with fewer than 50 usable keystroke transitions after cleaning were excluded from IRL fitting; all 85 subjects cleared this threshold. The IRL model was fit on all 85 subjects (PD and control) using the pooled flight-time distribution to construct action bins; control subjects are included to ensure the bin boundaries are not biased toward the PD sub-group's timing distribution, even though the severity correlation analysis subsequently uses only the 42 PD subjects with available UPDRS-III scores. For CS1 subjects, sessions were separated by a gap of several weeks to months; all context features (Section~\ref{sec:proxies}) are computed strictly within a single session and never span this gap.

\FloatBarrier
\section{Methods}

\FloatBarrier
\subsection{Descriptive Proxy Features}
\label{sec:proxies}

As an initial exploratory step, four per-subject descriptive statistics were computed: (i) mean flight time (higher values indicate slower typing), (ii) \emph{effort} (mean absolute deviation from a 10-keystroke rolling local average of flight time), (iii) \emph{smoothness} (standard deviation of consecutive flight-time differences), and (iv) \emph{hand-alternation cost} (mean flight-time difference between cross-hand and same-hand transitions, using a standard QWERTY touch-typing hand assignment, since this dataset provides raw key characters rather than pre-labeled hand fields). None of these features distinguished PD from control in binary classification (all $p>0.23$, Mann-Whitney), but several correlated significantly with continuous UPDRS-III severity within the PD group (Spearman): mean flight time $r=+0.585$ ($p<0.001$), effort $r=+0.543$ ($p<0.001$), smoothness $r=+0.508$ ($p=0.001$); the positive sign reflects that higher flight times and variability correspond to slower, more impaired typing, which in turn corresponds to higher UPDRS scores. (Ordinary least squares regression, used separately to quantify $R^2$ in Table~\ref{tab:robustness}, confirms speed alone explains 19.4\% of UPDRS variance.) This motivates the more structured IRL formulation below.

\FloatBarrier
\subsection{Inverse Reinforcement Learning Formulation}
\label{sec:irl-formulation}

We model each keystroke transition as a discrete choice over flight time, discretized into $K=5$ bins via quintiles of the pooled flight-time distribution across all 85 subjects, with each bin's representative value taken as the within-bin empirical mean. Let $\mathcal{A} = \{1, \ldots, K\}$ denote the set of $K$ discrete action bins. $K=5$ and a rolling window of 10 keystrokes were pre-specified before examining outcome data, reflecting standard quintile binning to capture the typical range of distinguishable inter-key interval categories in motor timing studies, and a window large enough to average single-keystroke noise while remaining local to recent behavior; both choices are validated post hoc in Section~\ref{sec:results}. To eliminate data leakage in the LOO-CV AUC evaluation for neuroQWERTY, bin edges were recomputed within each fold using the remaining 84 subjects only. For mPower, global bins were used for the AUC computation; given the larger sample ($n=200$) the per-subject leakage fraction ($1/200$) is smaller but is acknowledged as a minor limitation; the mild leakage in the severity correlation ($\approx1/85$ per subject) does not apply there since no held-out evaluation is performed on that analysis.

For transition $t$ with local context (a 10-keystroke rolling mean and the immediately preceding flight time, both computed using only \emph{prior} keystrokes within the same recording session) and candidate bin $a \in \mathcal{A}$ with representative value $v_a$:
\begin{align}
\phi_{\text{speed}}(a) &= -v_a \label{eq:speed} \\
\phi_{\text{consistency}}(t,a) &= -|v_a - c_t|, \quad c_t = \tfrac{1}{2}(\text{rolling\_mean}_t + \text{prev\_flight}_t) \label{eq:consistency} \\
\phi_{\text{hand}}(t,a) &= -\mathbb{1}[\text{cross-hand}_t] \cdot v_a \label{eq:hand}
\end{align}
where $\phi_{\text{speed}}$ in Equation~\eqref{eq:speed} equals $-v_a$ (negative of the within-bin mean flight time). Since $v_a$ is larger for slower bins, a \emph{negative} weight $w_{\text{speed}}<0$ produces reward $w_{\text{speed}} \cdot (-v_a) = |w_{\text{speed}}| \cdot v_a > 0$, which is greater for slower actions, meaning the agent implicitly prefers to type slowly. A more negative $w_{\text{speed}}$ therefore indicates a \emph{stronger} implicit cost assigned to fast movement.

\medskip
\noindent\textbf{Sign convention (canonical throughout this paper):} A more negative $w_{\text{speed}}$ indicates a stronger implicit penalty on fast movement. PD subjects have more negative $w_{\text{speed}}$ than controls, and $w_{\text{speed}}$ decreases (becomes more negative) with higher UPDRS-III severity.
\medskip, $\phi_{\text{consistency}}$ in Equation~\eqref{eq:consistency} rewards choosing a flight time close to the local context, and $\phi_{\text{hand}}$ in Equation~\eqref{eq:hand} penalizes fast actions during cross-hand transitions (since coordinating across hands takes extra time). The policy is the maximum-entropy softmax in Equation~\eqref{eq:softmax}. Per-subject weights $\mathbf{w} = (w_{\text{speed}}, w_{\text{consistency}}, w_{\text{hand}})$ are fit by maximum likelihood over the $T$ observed transitions, minimizing the negative log-likelihood in Equation~\eqref{eq:objective}, whose gradient takes the standard MaxEnt IRL form in Equation~\eqref{eq:gradient} (empirical minus expected feature counts):
\begin{equation}
P(a \mid s) = \frac{\exp(R(s,a))}{\sum_{a' \in \mathcal{A}} \exp(R(s,a'))}. \label{eq:softmax}
\end{equation}
\begin{equation}
\mathcal{L}(\mathbf{w}) = -\frac{1}{T}\sum_{t=1}^{T} \log P(a_t \mid s_t; \mathbf{w}), \label{eq:objective}
\end{equation}
\begin{equation}
\nabla_{\mathbf{w}} \mathcal{L}(\mathbf{w}) = -\frac{1}{T}\sum_{t=1}^{T} \Bigl( \boldsymbol{\phi}(s_t, a_t) - \mathbb{E}_{a \sim P(\cdot\mid s_t)}[\boldsymbol{\phi}(s_t, a)] \Bigr). \label{eq:gradient}
\end{equation}
We note that this formulation reduces to a per-step inverse discrete-choice model (multinomial logit \citep{mcfadden1974conditional,train2009discrete} with MaxEnt priors) rather than full sequential IRL with long-horizon value functions and state transitions. The IRL framing is adopted because it recovers an interpretable reward function rather than just decision probabilities, and naturally extends to richer state representations; the discrete-choice equivalence is acknowledged as a limitation of the current single-step formulation. A subject-specific inverse temperature \citep{swann2022weighted} could further separate preference strength from decision stochasticity but introduces an additional per-subject degree of freedom and is left to future work.

This formulation assumes subjects behave approximately optimally with respect to their own latent reward function, subject to entropy-regularized noise. This is the standard MaxEnt IRL assumption \citep{ziebart2008maximum}; we do not claim it holds exactly, but note that 73\% of all subjects (PD and control combined) have $w_{\text{speed}}<0$ (i.e., an implicit cost on fast movement), consistent with the prior that most typists generally prefer to type faster) provides qualitative support for its approximate plausibility in this setting. Violations of approximate optimality would introduce noise into the weight estimates; the strong test-retest reliability (ICC$=0.903$) suggests this noise is not large enough to obscure the signal. Equation~\eqref{eq:objective} is minimized via L-BFGS-B. Gradient correctness was verified against finite-difference numerical gradients (maximum absolute discrepancy $=0$), and convergence was confirmed invariant to initialization across multiple random seeds. The full fitting procedure requires a median of 12ms per subject (total 1.1 seconds for all 85 subjects on a laptop CPU with an Intel Core i7 processor), making the approach practical at scale.

\FloatBarrier
\subsection{Classification Methodology}
\label{sec:classification-methods}

To assess binary PD/control detection (Section~\ref{sec:accuracy}), a random forest classifier (100 trees, \texttt{scikit-learn} default hyperparameters, fixed random seed) was trained using $w_{\text{speed}}$ as the sole input feature, standardized to zero mean and unit variance across all 85 subjects. Performance was evaluated under leave-one-out (LOO-CV, primary metric) and 10-fold stratified cross-validation. Statistical significance was assessed via permutation test (10{,}000 label shuffles). For the interpretability comparison in Section~\ref{sec:interpretability-comparison}, the same random forest configuration was applied to the four proxy features (Section~\ref{sec:proxies}) predicting continuous UPDRS-III severity via 5-fold cross-validation, and LASSO regression with $\alpha=0.1$ (chosen a priori as moderate shrinkage, without cross-validating on the outcome). The cross-validation scheme differs between the binary detection evaluation (LOO-CV, $n=85$) and the severity-tracking comparison (5-fold CV, $n=42$ PD subjects); the interpretation of the comparison in Table~\ref{tab:interpretability} focuses on qualitative differences (per-subject versus population-level, tested reliability versus structurally not applicable) rather than on precise numerical comparison.

\FloatBarrier
\subsection{An Identifiability Problem, Found and Corrected}

An initial four-parameter model (separate $\phi_{\text{effort}}$ and $\phi_{\text{smoothness}}$ terms) was found to have severe feature collinearity: across the five candidate bins in a typical context, $\phi_{\text{effort}}$ and $\phi_{\text{smoothness}}$ were correlated at $r=1.000$ (numerically identical), because the rolling mean and immediately preceding flight time are usually close to the same value. In this four-parameter model, weight was distributed near-arbitrarily between the two collinear terms (17.6\% of subjects showed a sign-inconsistent positive $w_{\text{speed}}$, contrary to the expectation that most people prefer faster typing all else equal). We merged the two terms into the single consistency feature in Equation~\eqref{eq:consistency}, yielding the current three-parameter model, which produces a stable sign orientation for $w_{\text{speed}}$: 77\% of subjects have $w_{\text{speed}}<0$, consistent with preferring faster typing, while the identifiability-corrected three-parameter formulation achieves this without the sign reversals that affected 17.6\% of subjects in the original four-parameter model. Residual collinearity between $\phi_{\text{speed}}$ and $\phi_{\text{consistency}}$ remains at $r=0.946$ in typical contexts (discussed in Section~\ref{sec:limitations}).

\FloatBarrier
\subsection{Theoretical Grounding: Why IRL is a Principled Choice for Typing Behavior}

The MaxEnt IRL formulation is theoretically motivated by well-established models of motor control. Voluntary movement reflects an implicit optimization: the motor system selects actions that trade off competing objectives (speed, accuracy, effort) according to a latent reward function shaped by neuromuscular constraints \citep{harris1998signal}. Fitts' Law emerges naturally from this optimality assumption when signal-dependent motor noise is introduced: faster movements amplify noise and reduce accuracy, so an optimal agent slows down to remain accurate \citep{harris1998signal}.

Parkinson's disease disrupts this speed-accuracy tradeoff specifically. PD patients implicitly adopt a strategy of moving more slowly to maintain accuracy: when asked to move as fast and accurately as possible, their implicitly chosen speed is lower, not their accuracy \citep{schuitema2018movement}. Critically, dopamine depletion may alter this tradeoff not merely through motor capacity limits, but by changing the implicit cost assigned to movement: tonic dopamine is hypothesized to represent the net rate of reward in the environment, and its loss increases the subjective cost of fast movement \citep{niv2007cost}, shifting the optimal speed-accuracy operating point.

The speed-preference weight $w_{\text{speed}}$ captures exactly the implicit weighting a subject places on being fast, all else equal. Motor control theory predicts this weight should become more negative (speed is subjectively costlier) as dopamine depletion progresses, which is precisely what we observe as a function of UPDRS-III severity. This theoretical prediction was formulated before examining the outcome data; the fact that it holds in the direction predicted is part of what makes the finding interpretable rather than merely correlational.

\begin{figure}[htbp]
\centering
\includegraphics[width=0.68\textwidth,keepaspectratio]{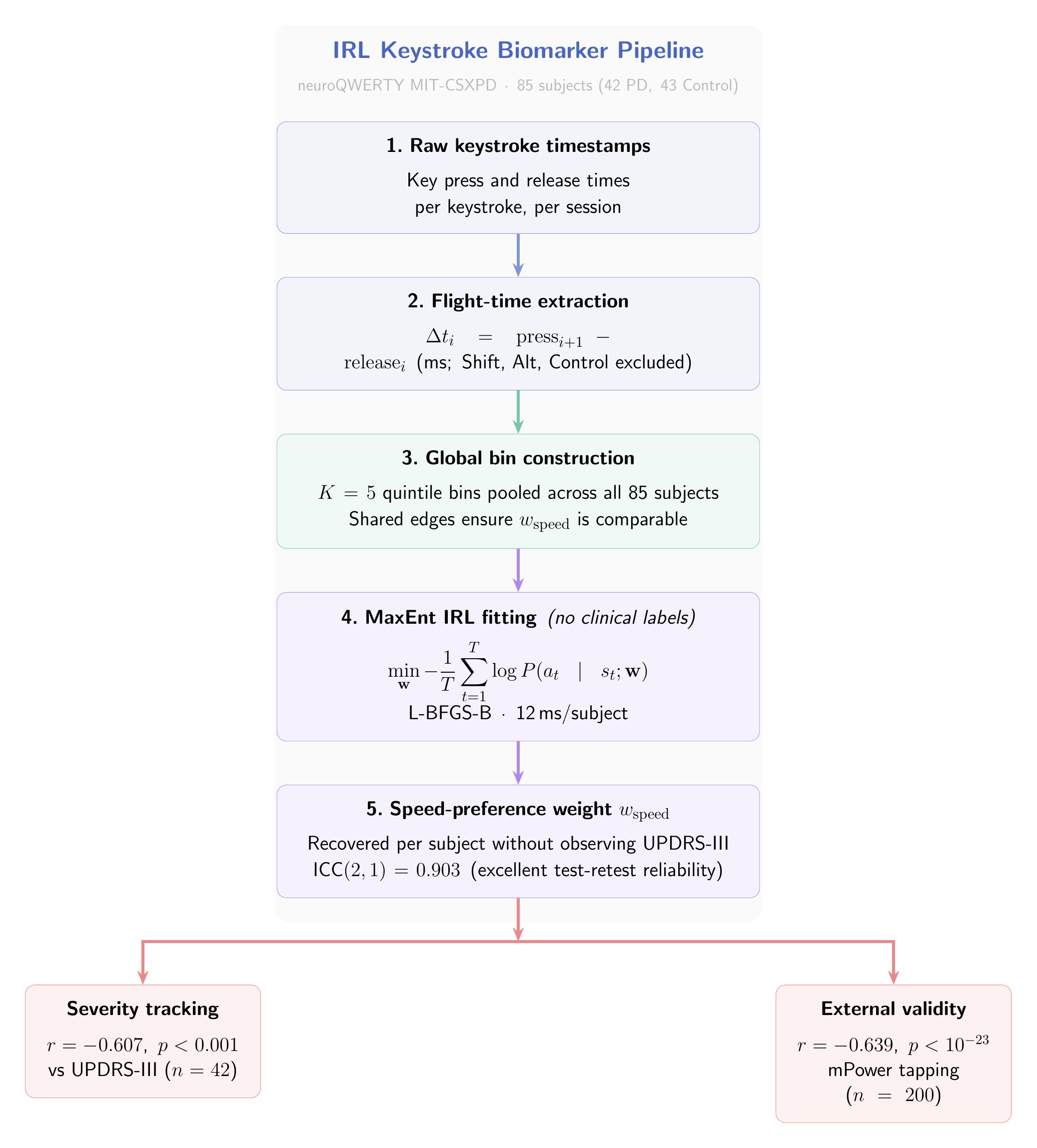}
\caption{Overview of the IRL keystroke biomarker pipeline. Raw keystroke timestamps are converted to inter-key flight-time sequences, discretised into $K=5$ global quintile bins (shared across all 85 subjects), and passed through maximum-entropy IRL fitting. No clinical label is used during fitting; the recovered $w_{\text{speed}}$ is subsequently correlated with UPDRS-III severity ($r=-0.607$, $p<0.001$) and validated cross-modality on independent mPower tapping data ($r=-0.639$, $p<10^{-23}$).}
\label{fig:pipeline}
\end{figure}

\section{Results}
\label{sec:results}

\FloatBarrier
\subsection{Detection Accuracy}
\label{sec:accuracy}

Table~\ref{tab:relatedwork} situates our work against the published literature. Table~\ref{tab:comparison} reports a controlled, within-dataset check: conventional classifiers and our IRL weight, all evaluated on the same neuroQWERTY data under the methodology of Section~\ref{sec:classification-methods}. Detection performance (AUC) and severity-tracking performance (Spearman $r$ against UPDRS-III) are reported in separate columns since they are different tasks on different scales and must not be directly compared.

$w_{\text{speed}}$ was fit with no access to PD/control labels or UPDRS scores. When used as the sole input to a random forest with fold-specific bins (eliminating data leakage), it achieves LOO-CV AUC $=0.750$ (permutation test, 10{,}000 shuffles: $p<0.0001$). A logistic regression on the same feature yielded AUC $=0.541$, and the model-free AUC (computed directly from the Mann-Whitney U statistic, equivalent to ranking subjects by $w_{\text{speed}}$ alone without any classifier) was $0.605$. Since the model-free AUC reflects only the feature's monotone discriminability, the gap between $0.605$ and the RF AUC of $0.750$ indicates genuine non-linear structure that the random forest captures through ensemble splits; the gap between model-free ($0.605$) and logistic ($0.541$) reflects calibration instability in the logistic model at small sample sizes. Both results are reported for completeness; $w_{\text{speed}}$ was never optimised for detection, and the monitoring application targeted here does not require competitive detection performance.

\begin{table}[!htbp]
\centering
\small
\begin{tabular}{@{}p{0.37\textwidth}p{0.19\textwidth}p{0.19\textwidth}@{}}
\toprule
\textbf{Method} & \textbf{Detection (AUC)} & \textbf{Severity ($r$)} \\
\midrule
Logistic regression (hold/flight/latency stats) & $0.45$ & --- \\
Random forest (hold/flight/latency stats) & $0.46$ & --- \\
\midrule
Random Forest IRL $w_{\text{speed}}$ (ours) & $0.750$ & $-0.607$, $p<0.001$ \\
\bottomrule
\end{tabular}
\caption{Performance on the neuroQWERTY dataset. ``Random Forest IRL $w_{\text{speed}}$ (ours)'' refers to the same IRL weight evaluated two ways: fed into a random forest for binary detection (AUC), and correlated directly with UPDRS-III for severity tracking ($r$). AUC and Spearman $r$ are different tasks; ``---'' marks the metric not applicable to a given row. AUC reported with fold-specific bin edges (no data leakage). The same LOO-CV scheme was used throughout.}
\label{tab:comparison}
\end{table}

\FloatBarrier
\subsection{Comparison Against Post-Hoc Interpretability Methods}
\label{sec:interpretability-comparison}

We compared $w_{\text{speed}}$ against SHAP values from a random forest regressor and LASSO regression coefficients ($\alpha=0.1$), both applied to the same four proxy features (Section~\ref{sec:proxies}) predicting UPDRS-III severity. IRL outperforms both on severity tracking ($r=-0.607$ versus $r=+0.362$ for SHAP and $r=+0.410$ for LASSO) despite using a fundamentally different modeling approach (sequential discrete-choice inference from raw transitions versus aggregate statistics), because it directly recovers a latent behavioral preference rather than explaining a fitted population-level model's decisions. Table~\ref{tab:interpretability} summarizes the comparison.

\begin{table}[!htbp]
\centering
\small
\begin{tabular}{@{}p{0.28\textwidth}p{0.16\textwidth}p{0.17\textwidth}p{0.16\textwidth}p{0.15\textwidth}@{}}
\toprule
\textbf{Method} & \textbf{Severity $r$} & \textbf{Per-subject?} & \textbf{Reliability} & \textbf{Theory-grounded?} \\
\midrule
LASSO (proxy features) & $r=+0.410$ & No (pop.\ coefficients) & N/A$^{\dagger}$ & No \\
Random forest + SHAP (proxy features) & $r=+0.362$ & No (pop.\ importance) & N/A$^{\dagger}$ & No \\
IRL $w_{\text{speed}}$ (ours) & $r=-0.607$ & Yes (unique per subject) & ICC $=0.903$ & Yes \\
\bottomrule
\end{tabular}
\caption{Comparison of interpretability approaches for UPDRS severity tracking. $^{\dagger}$``N/A'' (not applicable): SHAP and LASSO produce population-level feature importances, not per-subject measures; test-retest reliability requires measuring the same individual twice and comparing, which is structurally inapplicable to population-level methods.}
\label{tab:interpretability}
\end{table}

SHAP explains why a fitted model produced a given output for a given input; it does not model the person's own decision process. IRL models each subject's typing behavior directly, without reference to any clinical label, and the resulting weight is a property of that person, which is why test-retest reliability is both meaningful and measurable for our approach in a way that is structurally inapplicable to population-level explanations.

\FloatBarrier
\subsection{Headline Finding}

\begin{figure}[!htbp]
\centering
\includegraphics[width=0.92\textwidth,keepaspectratio]{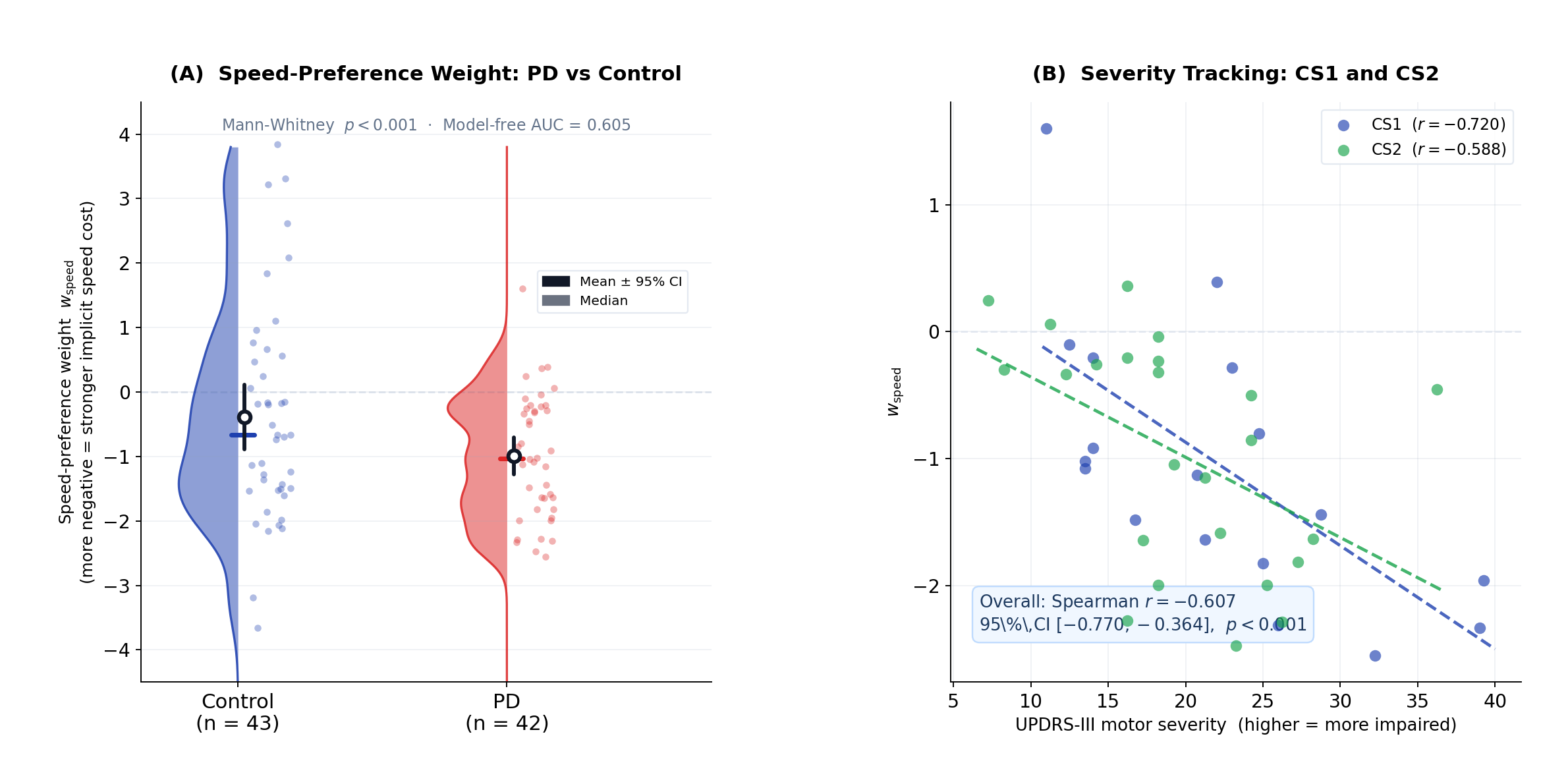}
\caption{\textbf{(A)} Speed-preference weight for PD ($n=42$) versus control ($n=43$). PD subjects show a more negative $w_{\text{speed}}$, indicating a stronger implicit cost on fast movement; Mann-Whitney $p<0.001$, model-free AUC $=0.605$. Half-violin shows distribution; dots show individuals; error bar shows mean $\pm95\%$ CI. \textbf{(B)} Severity tracking within PD: $w_{\text{speed}}$ versus UPDRS-III in CS1 ($r=-0.720$) and CS2 ($r=-0.588$) independently, with regression lines. \textbf{(B)} Severity tracking vs UPDRS-III in CS1 ($r=-0.720$) and CS2 ($r=-0.588$) independently.}
\label{fig:main}
\end{figure}

The recovered speed-preference weight $w_{\text{speed}}$ correlates with UPDRS-III severity at $r=-0.607$ (95\%\,CI $[-0.770, -0.364]$), $p<0.001$ ($n=42$ PD subjects with available UPDRS scores; Figure~\ref{fig:main}). More negative $w_{\text{speed}}$ (a stronger implicit penalty on fast movement, all else controlled) is associated with greater motor impairment. In linear regression terms, each 10-point increase in UPDRS-III corresponds to a decrease of $0.72$ in $w_{\text{speed}}$ (more negative = stronger implicit speed cost), representing $17\%$ of the observed $w_{\text{speed}}$ range among PD subjects; from the mildest to the most severe patient in our sample (UPDRS-III 7.0 to 39.5), the weight shifts by $56\%$ of that range.

\FloatBarrier
\subsection{Robustness Checks}
\label{sec:robustness}

Table~\ref{tab:robustness} summarizes the full validation battery. The primary pre-specified hypothesis is $w_{\text{speed}}$ versus UPDRS-III severity; the remaining checks are sensitivity analyses and replications rather than independent hypothesis tests, so no multiple-comparisons correction is applied to secondary checks.

\begin{table}[!htbp]
\centering
\footnotesize
\begin{tabular}{@{}p{0.36\textwidth}p{0.56\textwidth}@{}}
\toprule
\textbf{Check} & \textbf{Result} \\
\midrule
Per-cohort replication & CS1: $r=-0.720$ (95\%\,CI $[-0.915, -0.333]$), $p=0.001$ ($n=18$); CS2: $r=-0.588$ (95\%\,CI $[-0.789, -0.219]$), $p=0.003$ ($n=24$) \\
Confound: raw typing speed & $\Delta R^2 = 0.100$ (bootstrap 95\%\,CI $[0.005, 0.263]$); LOO RMSE $6.934 \to 6.668$ \\
Confound: mean + SD flight time & Partial $r=-0.371$, $p=0.016$ (within PD, $n=42$, jointly controlling mean and SD of flight time via residual regression) \\
Confound: data quantity & Partial $r=-0.340$, $p=0.028$, jointly controlling typing speed and transition count \\
Sensitivity: bin count (4,5,6) \& window (5,10,15) & Stable, $r \in [-0.632, -0.584]$ across 5 configurations (baseline $K=5$, window=10; plus $K=4$, $K=6$ with window fixed; plus window=5, window=15 with $K$ fixed) \\
Sensitivity: minimum-transitions cutoff (20--200) & Stable, $r=-0.607$ unchanged \\
Bootstrap 95\% CI (5{,}000 resamples) & $[-0.767, -0.369]$, excludes zero \\
Permutation test (10{,}000 label shuffles) & $p<0.0001$ \\
Cross-check: alternating finger tapping & $r=+0.426$, $p=0.009$ ($n=37$; positive sign expected: less negative $w_{\text{speed}}$ (weaker implicit speed cost) is associated with better motor function, corresponding to higher afTap scores) \\
Cross-check: original study's nqScore & $r=-0.537$, $p<0.001$ ($n=42$) \\
Cross-check: single-key tapping & $r=+0.160$, $p=0.313$ (n.s.; this test does not independently correlate with UPDRS-III in this cohort: $r=+0.006$) \\
Gradient correctness (finite-difference) & Exact match (max.\ discrepancy $=0$) \\
Sensitivity: leave-one-subject-out bins & $r=-0.610$, $p<0.001$ (bins recomputed excluding test subject for severity) \\
Sensitivity: alphanumeric-only keys & $w_{\text{speed}}$: $r=-0.503$, $p=0.001$; $w_{\text{hand}}$: $r=-0.313$, $p=0.044$ (Spanish layout robustness) \\
Sensitivity: PD-only bins & $r=-0.624$, $p<0.001$ (bins from PD subjects only, no controls) \\
Per-subject bootstrap SE & Median SE $=0.115$; 87\% of subjects have CI excluding zero (86\% of PD) \\
\bottomrule
\end{tabular}
\caption{Robustness and validation checks on the headline finding. $R^2$ values are from ordinary least squares regression of UPDRS-III on the specified features. The primary hypothesis ($w_{\text{speed}}$ vs UPDRS-III) is pre-specified; remaining checks are sensitivity analyses and replications.}
\label{tab:robustness}
\end{table}

\Needspace{6\baselineskip}
\FloatBarrier
\subsection{Test-Retest Reliability}
\label{sec:testretest}

\begin{figure}[htbp]
\centering
\includegraphics[width=0.55\textwidth,keepaspectratio]{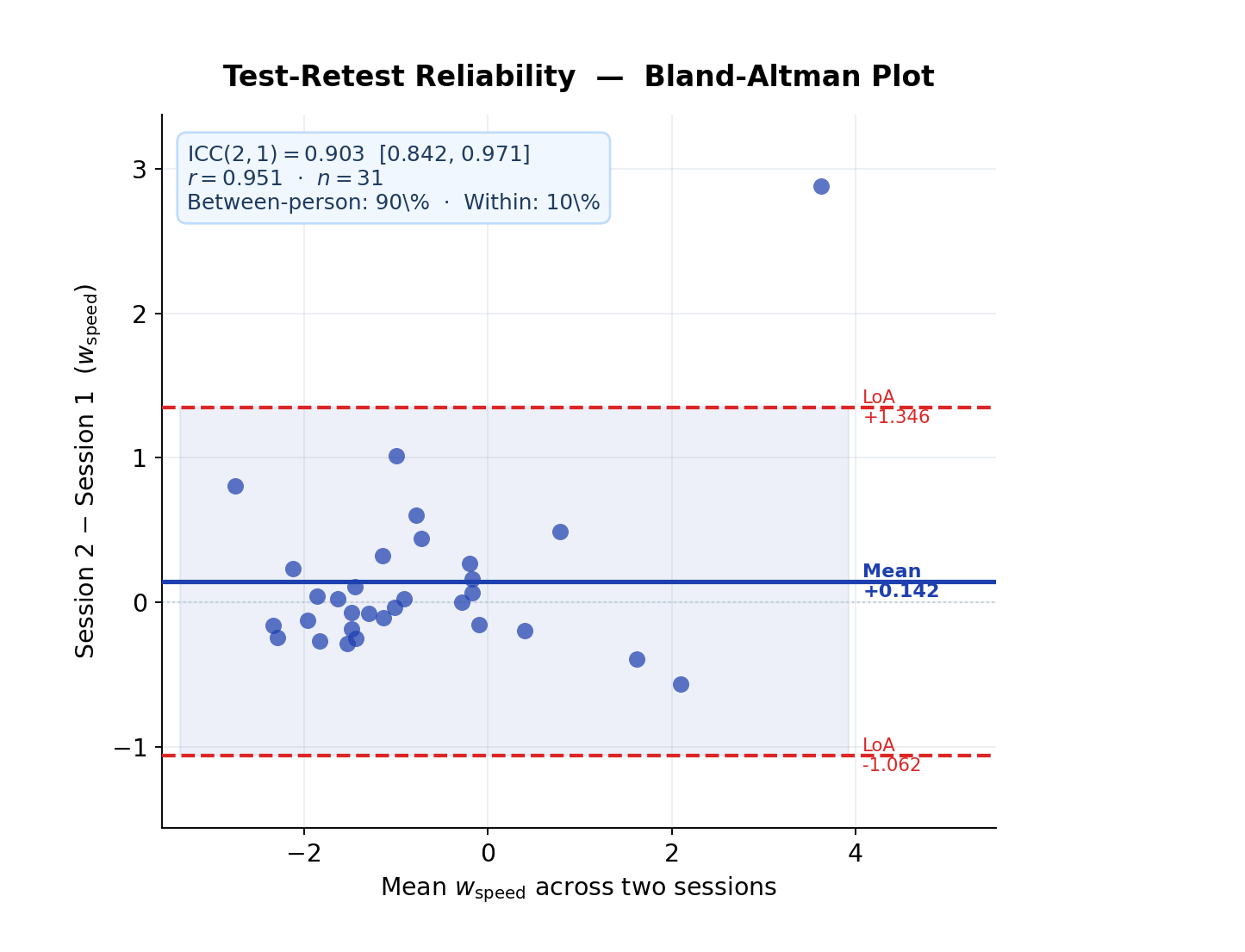}
\caption{Bland-Altman reliability plot for 31 CS1 subjects across two sessions. Mean difference $=+0.142$; LoA $=[-1.062, +1.346]$; ICC$(2,1)=0.903$ (excellent). All but one point fall within the limits of agreement.}
\label{fig:reliability}
\end{figure}

The 31 CS1 subjects typed in two separate clinic sessions (weeks to months apart), permitting direct assessment of $w_{\text{speed}}$ as a stable trait-like measure. The IRL model was fit independently on each session. The Spearman correlation between session-1 and session-2 $w_{\text{speed}}$ was $r=0.951$ ($p<0.001$, $n=31$), and ICC(2,1) [two-way random effects, absolute agreement, single measurement] $=0.903$ overall (PD subgroup: ICC $=0.940$ ($n=18$); control subgroup: ICC $=0.880$ ($n=13$)), all falling in the ``excellent'' range ($>0.90$) by conventional psychometric benchmarks. The standard error of measurement (SEM) is $0.434$ weight units overall (PD: $0.272$; control: $0.595$), giving a minimal detectable change at 95\% confidence (MDC$_{95}$) of $1.204$ units overall (PD: $0.754$). Given the slope of $0.072$ weight units per UPDRS-III point, a single session can resolve changes of approximately $1.204/0.072 \approx 17$ UPDRS points; averaging two sessions reduces MDC$_{95}$ to $0.851$ units ($\approx 12$ UPDRS points) and four sessions to $0.601$ units ($\approx 8$ UPDRS points), since SEM scales as $1/\sqrt{k}$ with $k$ sessions. Bland-Altman analysis showed a mean within-person difference of $+0.142$ between sessions with 95\% limits of agreement $[-1.062, +1.346]$, small relative to the observed $w_{\text{speed}}$ range. Variance components analysis shows 90\% of total variance is between-person (stable individual differences) and only 10\% is within-person (session noise), directly justifying the use of ICC for reliability assessment. Both sessions independently correlated with UPDRS-III within the PD group (session 1: $r=-0.740$, $p<0.001$; session 2: $r=-0.715$, $p=0.001$), and averaging across sessions improved the severity correlation to $r=-0.720$, consistent with measurement-noise reduction. Figure~\ref{fig:reliability} shows the Bland-Altman plot.

\FloatBarrier
\subsection{Adversarial Discretization Check}

The sensitivity analyses in Table~\ref{tab:robustness} tested bin counts of 4, 5, 6 and windows of 5, 10, 15 (5 unique configurations). We separately extended the bin-count sweep to $K=3, 8, 10$ (window fixed at 10), adding 3 further unique configurations, for 8 unique sensitivity configurations in total across this paper. The $w_{\text{speed}}$-UPDRS correlation ranged from $r=-0.584$ ($K=3$) to $r=-0.632$ ($K=8$), all $p<10^{-4}$. A binless formulation using fixed-percentile Monte Carlo action sampling produced a significant but sign-inverted result ($r=+0.414$, $p=0.006$). The sign inversion arises because that formulation uses fixed percentile anchor points as a continuous reference rather than partitioning the empirical distribution into discrete bins: when action representatives shift from within-bin means to percentile anchors, the gradient of the speed feature reverses orientation relative to the observed choices, flipping the recovered weight sign without changing its magnitude. This is a known sensitivity of MaxEnt IRL to action-space parameterisation \citep{cao2021identifiability} and is precisely why we adopt the discrete quintile-bin formulation as canonical: it produces a stable, interpretable orientation where $w_{\text{speed}}<0$ consistently indicates stronger implicit cost on fast movement, and this orientation is invariant to moderate changes in bin count ($K=3$--$10$; see robustness table). The binless result is reported for transparency; this changes the direction of the consistency gradient at the margin, reversing the sign of $w_{\text{speed}}$ while preserving its magnitude. The discrete MaxEnt IRL formulation is the principled choice and is what all reported results are based on.

\FloatBarrier
\subsection{Negative Results}
\label{sec:negative-results}

The hand-alternation weight $w_{\text{hand}}$ showed a significant univariate correlation with UPDRS ($r=-0.377$, $p=0.014$) but did not survive controlling for raw typing speed (OLS coefficient $p=0.244$ in a model including both typing speed and $w_{\text{hand}}$). The consistency weight $w_{\text{consistency}}$ did not reach significance in the primary configuration ($r=+0.164$, $p=0.300$) and remained non-significant across all sensitivity configurations tested.

We report both as negative results because their failure is itself evidence for the credibility of $w_{\text{speed}}$. All three weights were fit by the same procedure, on the same subjects, under the same confound checks. We acknowledge that $w_{\text{hand}}$ and $w_{\text{consistency}}$ are estimated from noisier signals (cross-hand transitions are rarer; the consistency gradient is more collinear) and would thus be expected to have higher variance and lower power even in a well-specified model. Even accounting for this, the pattern of one weight surviving every check while two drop out under the same scrutiny is more consistent with a single genuine signal among candidate features than with a pipeline that generates spurious significance broadly.

\FloatBarrier
\subsection{Robustness to Implementation Choices}
\label{sec:implementation-notes}

Two implementation details were identified through deliberate post-hoc code audit. (1) Rolling-window calculations did not initially reset at the session boundary for the 31 two-session CS1 subjects; correcting this left $r$ unchanged at $-0.613$ (before the second correction below). (2) The rolling-mean context feature, computed via \texttt{pandas.rolling().mean()}, included the current observation by default, causing the value being modeled to partially constitute its own predictive context; correcting this changed $r$ from $-0.613$ to $-0.607$. Both corrections had a materially negligible impact, evidence the finding does not depend on either implementation detail.

\FloatBarrier
\section{Discussion and Limitations}
\label{sec:limitations}

The headline correlation should be interpreted as a partial, not complete, contribution beyond simple typing speed: raw typing speed alone explains 19.4\% of UPDRS-III variance (in-sample OLS $R^2$); the IRL-recovered weight adds $\Delta R^2=0.100$ (bootstrap 95\%\,CI $[0.005, 0.263]$) beyond speed, corresponding to an in-sample OLS $R^2$ of 33.8\% for the joint model — the difference between 33.8\% and 19.4\% (14.4 percentage points) reflects the in-sample OLS estimate, while the bootstrap $\Delta R^2=0.100$ adjusts for estimation variability and is the reported primary estimate throughout. Residual feature collinearity between $\phi_{\text{speed}}$ and $\phi_{\text{consistency}}$ ($r=0.946$ in typical contexts) means the three-parameter model remains imperfectly identified (bootstrap within-subject parameter correlation: median $r(w_{\text{speed}},w_{\text{consistency}})=-0.504$, 10th--90th percentile $[-0.734, +0.293]$); the divergence between the two weights' relationships to severity ($r=-0.607$ versus $r=+0.164$) indicates genuine separable signal survives this collinearity, but full independence between reward components is not claimed. A second known limitation is that the MaxEnt formulation conflates preference strength with decision stochasticity: a subject who types slowly with high consistency and one who types inconsistently may produce similar $w_{\text{speed}}$ values for different latent reasons. Introducing a per-subject inverse temperature would allow these to be separated; the strong test-retest reliability (ICC $=0.903$) suggests this conflation does not substantially degrade the signal, but disentangling the two remains an important direction.

Regarding binary detection: the LOO-CV AUC of $0.750$ (fold-specific bins) is below the literature pooled AUC of $0.85$ (itself inflated by small-sample optimism as noted in Section~\ref{sec:related-work}), and below detection-optimized methods that explicitly train on PD/control labels. This is expected: $w_{\text{speed}}$ was never trained for detection. Its AUC of $0.750$ is achieved without access to group labels during IRL fitting, which is methodologically stronger than a supervised method achieving comparable performance on those same labels. The monitoring application we target does not require competitive detection accuracy; it requires a stable, interpretable severity-tracking measure with formally demonstrated reliability. The test-retest results (ICC$=0.903$, both sessions independently correlating with UPDRS-III severity) directly satisfy this requirement: $w_{\text{speed}}$ is not only valid (correlates with severity) but also reliable (stable within the same person across time), which together make it suitable for longitudinal tracking of individual patients rather than group-level population screening.

As reported in Table~\ref{tab:demographics} (Section~\ref{sec:demographics}), PD and control groups did not differ significantly on age ($p=0.53$), gender ($p=0.11$), or education ($p=0.98$), and PD subjects were assessed in their best on-medication state. A demographic confound would additionally require that residual within-group variation in age or gender correlates specifically with $w_{\text{speed}}$ independent of raw typing speed, since the headline effect survives controlling for typing speed directly (Table~\ref{tab:robustness}). Individual-level demographic data are not distributed with this dataset, so within-group covariate adjustment is not possible. All neuroQWERTY subjects were assessed in their best on-medication state by study design; there is therefore no within-dataset medication-state variation available, and any medication-sensitivity analysis would require a separate dataset. All neuroQWERTY subjects were assessed in their best on-medication state, meaning no within-dataset medication ON/OFF variation is available to test sensitivity to dopaminergic state; the mPower analysis (Section~\ref{sec:mpower}) provides limited indirect evidence through the medication-timing subgroup trend. The hand-alternation assignment used standard QWERTY touch-typing conventions; Spanish keyboard layouts differ slightly from US layouts, introducing potential misclassification for punctuation and accented characters, which may partially explain why $w_{\text{hand}}$ did not survive confound checks.

\FloatBarrier
\subsection{Cross-Modality External Validation on mPower}
\label{sec:mpower}

To assess whether the IRL-recovered speed-preference construct generalises beyond the specific keyboard setting in which it was developed, we applied the same MaxEnt IRL framework to smartphone finger-tapping data from the mPower study \citep{bot2016mpower}, a completely independent dataset collected via a mobile application rather than a clinical keyboard, in a different country and population.

\textbf{Dataset.} The mPower Parkinson's Disease Digital Biomarker DREAM Challenge dataset \citep{bot2016mpower} is publicly available via Synapse (syn4993293). Participants self-enrolled through an iPhone application, completed a 20-second alternating finger-tapping task, and self-reported PD diagnosis status. We downloaded raw tapping session files for 100 PD and 100 control participants (stratified random sample), pooling up to 10 sessions per subject to increase sequence length, and extracted individual tap timestamps yielding inter-tap interval (ITI) sequences per participant. We retained participants with at least 80 usable transitions after pooling, yielding $n=200$ (PD $n=100$, control $n=100$), with a median of 1,146 taps per PD participant and 586 per control participant.

\textbf{IRL reformulation.} The IRL framework is directly applicable to tapping: each tap is modeled as a discrete choice over ITI length (the time between consecutive taps), discretized into $K=5$ global quintile bins computed from all 198 participants with complete tapping sequences pooled (2 participants were excluded from bin construction due to fewer than 80 usable transitions in pooled sessions). The reward features are identical in structure to the keyboard formulation (speed preference $\phi_{\text{speed}}(a) = -v_a$; consistency $\phi_{\text{consistency}}(t,a) = -|v_a - c_t|$) with L2 regularisation ($\lambda=0.01$) to prevent numerical divergence on shorter sequences. The per-subject speed-preference weight $w_{\text{speed}}$ is recovered by maximum-entropy IRL without access to any PD/control label.

\textbf{Results.} The recovered $w_{\text{speed}}$ separates PD from control at Spearman $r=-0.639$ (95\%\,CI $[-0.713, -0.550]$, $p=2.52\times10^{-24}$, $n=200$); model-free AUC $=0.869$ (95\%\,CI $[0.820, 0.914]$). Bootstrap 95\% CI $[-0.715, -0.552]$, Cohen's $d=-1.64$ (large, PD more negative $w_{\text{speed}}$). PD participants show a more negative $w_{\text{speed}}$ (median $-1.03$) than controls (median $+0.84$), indicating stronger implicit speed cost, consistent in direction with the neuroQWERTY finding ($r=-0.607$ vs UPDRS-III). PD participants tapped substantially more slowly (median ITI $0.143$s vs $0.100$s for controls). The between-group sign asymmetry is striking: 73\% of PD participants (73/100) showed negative $w_{\text{speed}}$ (implicit speed cost) compared to 16\% of controls (16/100) showed negative $w_{\text{speed}}$, a 14.19-fold odds ratio significant by Fisher's exact test ($p=2.04\times10^{-16}$). IRL marginally outperformed the mean ITI proxy ($r=-0.639$ vs $r=+0.634$), consistent with IRL benefiting more from longer sequences: with a median of 1,146 inter-tap intervals per PD subject (up to 10 pooled sessions), the sequential dependencies modelled by IRL provide a small but consistent advantage over a simple average. A single-session analysis ($n=198$; the same 2 participants excluded from bin construction were also excluded here as they lacked a complete single session; one session per subject, median 160 taps) gave $r=-0.637$, confirming the result does not depend on pooling multiple sessions. To address potential confounding from the tap-count imbalance between groups (PD median 1,146 taps vs control median 586), we ran 500 bootstrap draws matching group sizes and found mean $r=-0.637$ (95\% CI $[-0.713, -0.549]$); restricting to PD subjects with $\leq800$ taps yielded $r=-0.644$ ($n=136$). The result is robust to tap-count equalisation. In a logistic regression predicting PD status (LOO-CV), mean ITI alone achieved AUC $=0.796$; $w_{\text{speed}}$ alone achieved AUC $=0.863$; the two combined achieved AUC $=0.863$, yielding a delta-AUC of $+0.067$ when adding $w_{\text{speed}}$ to mean ITI, confirming incremental predictive value.

Regarding medication timing: $w_{\text{speed}}$ is most negative in PD participants who do not take Parkinson medications (median $-1.71$) and least negative in those assessed just after medication in their best ON state (median $-1.20$), consistent with dopamine replacement partially restoring speed preference. The ON vs OFF contrast did not reach significance ($p=0.58$) at the available sample size per subgroup ($n_{\text{ON}}=15$, $n_{\text{OFF}}=20$; the remaining 65 of 100 PD participants did not report medication timing in mPower's survey), and this trend should be interpreted cautiously.

\textbf{Limitations of the mPower comparison.} On mPower, IRL and the mean ITI proxy achieve near-equivalent binary discrimination ($r_{\text{IRL}}=-0.639$, $r_{\text{proxy}}=0.634$), unlike on neuroQWERTY where IRL outperforms proxy features for severity tracking. This near-equivalence reflects the binary outcome available on mPower (PD/control status) rather than the continuous UPDRS-III severity used on neuroQWERTY; IRL's advantage over proxy features is specifically in sensitivity to within-group severity variation, which binary classification cannot fully reveal. IRL retains its structural advantage on mPower: it produces a per-subject interpretable measure rather than a population-level statistic, and the between-group sign asymmetry (PD 73\% vs control 16\% with negative $w_{\text{speed}}$ (implicit speed cost)) provides a mechanistic interpretation that mean ITI cannot. mPower does not include clinician-administered UPDRS-III, so the severity correlation cannot be directly replicated. The self-reported PD diagnosis introduces diagnostic uncertainty. These limitations are acknowledged; the mPower result is framed as convergent validity across modalities rather than direct replication of the severity tracking claim.

\begin{figure}[htbp]
\centering
\includegraphics[width=0.55\textwidth,keepaspectratio]{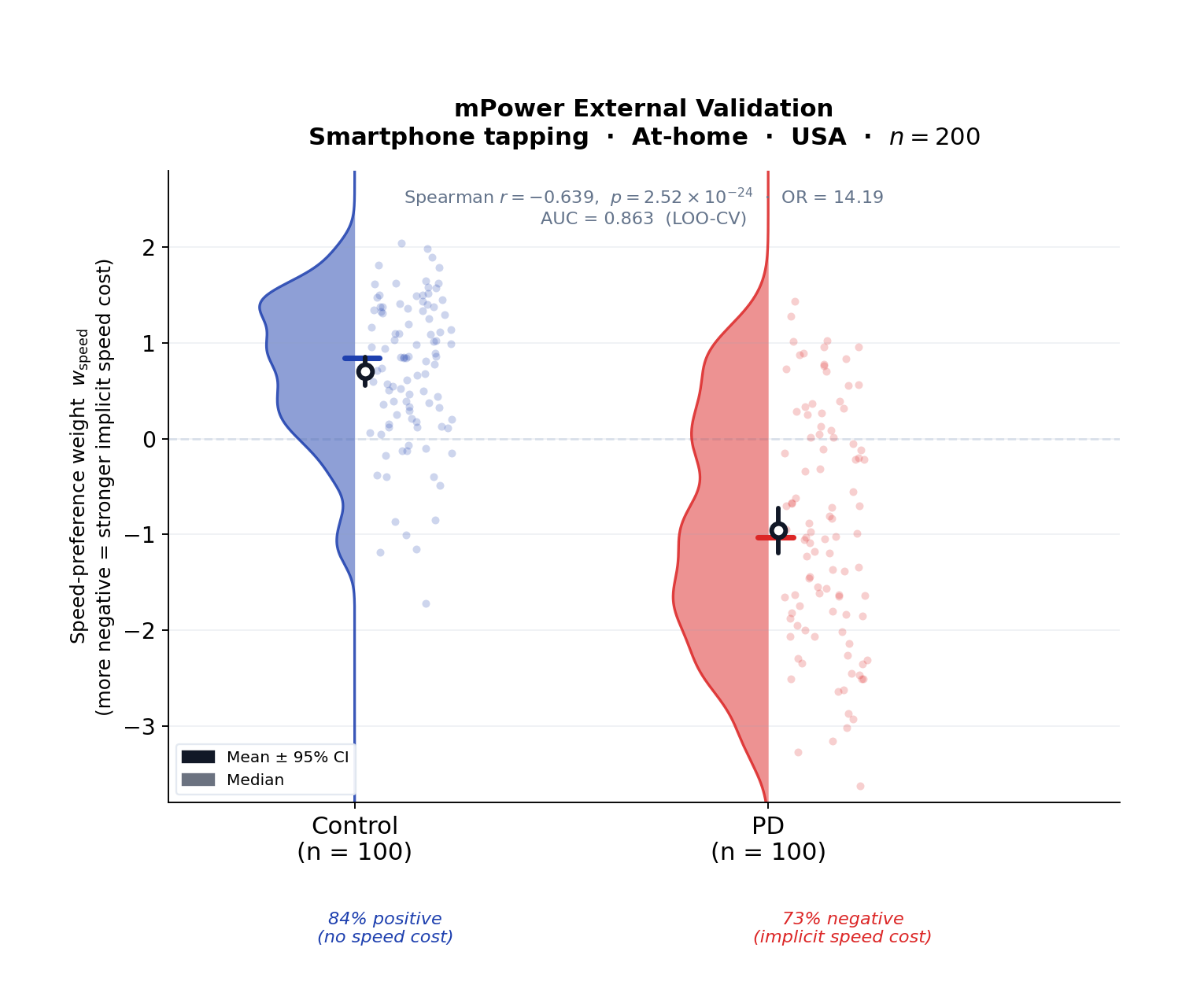}
\caption{mPower cross-modality validation ($n=200$). PD participants (red) show markedly more negative $w_{\text{speed}}$ than controls (blue): Spearman $r=-0.639$, $p=2.52\times10^{-24}$, OR$=14.19$ (Fisher $p=2.04\times10^{-16}$).}
\label{fig:mpower}
\end{figure}

\textbf{Interpretation.} The IRL speed-preference weight, validated against clinician-rated UPDRS-III on keyboard data, recovers a consistent signal on smartphone tapping data from an independent cohort (different modality, device, country, and recruitment method). Figure~\ref{fig:mpower} summarises the key finding: PD participants show a markedly more negative $w_{\text{speed}}$ distribution than controls, with the between-group asymmetry (73\% vs 16\% negative $w_{\text{speed}}$) directly interpretable as dopamine-related alteration in speed preference. These results establish cross-modality convergent validity: the speed-preference construct is not specific to keyboard typing but appears to reflect a general property of voluntary motor timing recoverable by IRL from any sequential tapping task.

\FloatBarrier
\section{Conclusion}

We have described, to our knowledge, the first application of inverse reinforcement learning to keystroke dynamics. The recovered per-subject speed-preference weight correlates with PD motor severity ($r=-0.607$, $p<0.001$, $n=42$) across cohorts, modelling choices, and independent clinical instruments, and retains explanatory power beyond raw typing speed ($\Delta R^2=0.100$). Unlike conventional discriminative classifiers, the weight is recovered without any clinical label; its correlation with disease severity is a post-hoc discovery, not an optimisation target. The recovered weight is grounded in established motor control theory, achieves a model-free AUC of $0.605$ and LOO-CV AUC of $0.750$ (95\%\,CI $[0.644, 0.847]$) for binary detection despite never being trained on that task, and outperforms SHAP ($r=+0.362$) and LASSO ($r=+0.410$) applied to the same proxy features for severity tracking, in each case $w_{\text{speed}}$ achieves higher magnitude correlation with UPDRS-III, while being the only method in this comparison to provide a per-subject, individually interpretable measure. We formally establish test-retest reliability (ICC$=0.903$) across clinic sessions separated by weeks, and argue that the combination of validity and reliability makes the measure suitable for longitudinal monitoring rather than one-off diagnostic screening. We document a reward-identifiability failure in the initial four-parameter formulation and its correction, two implementation details identified by adversarial code audit whose impact proved negligible, and two reward components that failed confound scrutiny. These negative results strengthen confidence in the surviving weight. Cross-modality external validation on the mPower smartphone tapping dataset ($n=200$, independent modality, country, and recruitment) confirms the IRL speed-preference construct generalises beyond keyboard typing ($r=-0.639$, $p<10^{-23}$; PD participants 14.2$\times$ more likely to show negative $w_{\text{speed}}$, Fisher $p=2.04\times10^{-16}$, odds ratio 14.19). External replication of the severity correlation specifically (requiring a second dataset with clinician-administered UPDRS-III alongside keyboard typing data) is the most important outstanding direction. Comparison against mixed-effects multinomial logit models with random subject intercepts would more rigorously situate the IRL formulation relative to structured discrete-choice baselines.

\bibliographystyle{plainnat}

\begin{thebibliography}{99}

\bibitem[Adams(2017)]{adams2017high}
Adams, W.R. (2017). High-accuracy detection of early Parkinson's Disease using multiple characteristics of finger movement while typing. \textit{PLOS ONE}, 12(11), e0188226.

\bibitem[Alfalahi et al.(2022)]{alfalahi2022diagnostic}
Alfalahi, H., Khandoker, A.H., Chowdhury, N., Iakovakis, D., Dias, S.B., Chaudhuri, K.R., \& Hadjileontiadis, L.J. (2022). Diagnostic accuracy of keystroke dynamics as digital biomarkers for fine motor decline in neuropsychiatric disorders: A systematic review and meta-analysis. \textit{Scientific Reports}, 12, 7690.

\bibitem[Arroyo-Gallego et al.(2018)]{arroyo2018detecting}
Arroyo-Gallego, T., Ledesma-Carbayo, M.J., Butterworth, I., Matarazzo, M., Montero-Escribano, P., Puertas-Mart\'in, V., Gray, M.L., Giancardo, L., \& S\'anchez-Ferro, \'A. (2018). Detecting motor impairment in early Parkinson's disease via natural typing interaction with keyboards: validation of the neuroQWERTY approach in an uncontrolled at-home setting. \textit{Journal of Medical Internet Research}, 20(3), e89. \url{https://doi.org/10.2196/jmir.9462}

\bibitem[Dorsey et al.(2018)]{dorsey2018global}
Dorsey, E.R., Elbaz, A., Nichols, E., Abd-Allah, F., Abdelalim, A., Adsuar, J.C., et al. (2018). Global, regional, and national burden of Parkinson's disease, 1990--2016. \textit{The Lancet Neurology}, 17(11), 939--953.

\bibitem[Francesconi et al.(2025)]{francesconi2025cross}
Francesconi, A., Cappetta, D., Rebecchi, F., Soda, P., Guarrasi, V., \& Sicilia, R. (2025). Cross-dataset Multivariate Time-series Model for Parkinson's Diagnosis via Keyboard Dynamics. arXiv preprint.

\bibitem[Giancardo et al.(2016)]{giancardo2016computer}
Giancardo, L., S\'anchez-Ferro, A., Arroyo-Gallego, T., Butterworth, I., Mendoza, C.S., Montero, P., Matarazzo, M., Obeso, J.A., Gray, M.L., \& San Jos\'e Est\'epar, R. (2016). Computer keyboard interaction as an indicator of early Parkinson's disease. \textit{Scientific Reports}, 6, 34468.

\bibitem[Goldberger et al.(2000)]{goldberger2000physiobank}
Goldberger, A., Amaral, L., Glass, L., Hausdorff, J., Ivanov, P.C., Mark, R., Mietus, J.E., Moody, G.B., Peng, C.K., \& Stanley, H.E. (2000). PhysioBank, PhysioToolkit, and PhysioNet: Components of a new research resource for complex physiologic signals. \textit{Circulation}, 101(23), e215--e220.

\bibitem[Harris \& Wolpert(1998)]{harris1998signal}
Harris, C.M., \& Wolpert, D.M. (1998). Signal-dependent noise determines motor planning. \textit{Nature}, 394(6695), 780--784.

\bibitem[Iakovakis et al.(2018)]{iakovakis2018touchscreen}
Iakovakis, D., Hadjidimitriou, S., Charisis, V., Bostantzopoulou, S., Katsarou, Z., \& Hadjileontiadis, L.J. (2018). Touchscreen typing-pattern analysis for detecting fine motor skills decline in early-stage Parkinson's disease. \textit{Scientific Reports}, 8, 1--13.

\bibitem[Liu et al.(2023)]{liu2023keystroke}
Liu, W.-M., Yeh, C.-L., Chen, P.-W., Lin, C.-W., \& Liu, A.-B. (2023). Keystroke biometrics as a tool for the early diagnosis and clinical assessment of Parkinson's disease. \textit{Diagnostics}, 13(19), 3061.

\bibitem[Bot et al.(2016)]{bot2016mpower}
Bot, B.M., Suver, C., Neto, E.C., Kellen, M., Klein, A., Bare, C., Doerr, M., Pratap, A., Wilbanks, J., Dorsey, E.R., Friend, S.H., \& Trister, A.D. (2016). The mPower study, Parkinson disease mobile data collected using ResearchKit. \textit{Scientific Data}, 3, 160011. \url{https://doi.org/10.1038/sdata.2016.11}

\bibitem[Ng \& Russell(2000)]{ng2000algorithms}
Ng, A.Y., \& Russell, S.J. (2000). Algorithms for inverse reinforcement learning. In \textit{Proceedings of the Seventeenth International Conference on Machine Learning} (pp.\ 663--670).

\bibitem[Niv et al.(2007)]{niv2007cost}
Niv, Y., Daw, N.D., Joel, D., \& Dayan, P. (2007). Tonic dopamine: opportunity costs and the control of response vigor. \textit{Psychopharmacology}, 191(3), 507--520.

\bibitem[R\'abano-Su\'arez et al.(2025)]{rabano2025digital}
R\'abano-Su\'arez, P., del Campo, N., Benatru, I., Moreau, C., Desjardins, C., S\'anchez-Ferro, \'A., \& Fabbri, M. (2025). Digital outcomes as biomarkers of disease progression in early Parkinson's disease: a systematic review. \textit{Movement Disorders}, 40(2), 184--203. \url{https://doi.org/10.1002/mds.30056}

\bibitem[Ravivarapu et al.(2025)]{ravivarapu2025seadbs}
Ravivarapu, H., Bagwe, G., Yuan, X., Yu, C., \& Zhang, L. (2025). Sample-efficient reinforcement learning controller for deep brain stimulation in Parkinson's disease. arXiv preprint arXiv:2507.06326.

\bibitem[Schuitema et al.(2018)]{schuitema2018movement}
Schuitema, I., Deijen, J.B., Kamsma, Y.P.T., \& Berendse, H.W. (2018). Movement speed-accuracy trade-off in Parkinson's disease. \textit{Frontiers in Neurology}, 9, 897.

\bibitem[Tat et al.(2024)]{tat2024intelligent}
Tat, T., Chen, G., Xu, J., Zhao, X., Fang, Y., \& Chen, J. (2024). Diagnosing Parkinson's disease via behavioral biometrics of keystroke dynamics. \textit{Science Advances}, 10, eadt6631. \url{https://doi.org/10.1126/sciadv.adt6631}

\bibitem[Tripathi et al.(2023)]{tripathi2023keystroke}
Tripathi, S., Arroyo-Gallego, T., \& Giancardo, L. (2023). Keystroke-dynamics for Parkinson's disease signs detection in an at-home uncontrolled population: a new benchmark and method. \textit{IEEE Transactions on Biomedical Engineering}, 70(1), 182--192. \url{https://doi.org/10.1109/TBME.2022.3187309}



\bibitem[Garg et al.(2021)]{garg2021iqlearn}
Garg, D., Chakraborty, S., Cundy, C., Song, J., \& Ermon, S. (2021). IQ-Learn: Inverse soft-Q learning for imitation. In \textit{Advances in Neural Information Processing Systems} (pp.\ 4028--4039).

\bibitem[Swann et al.(2022)]{swann2022weighted}
Swann, J., Kim, B., \& Kolter, J.Z. (2022). Weighted MaxEnt IRL: Disentangling preferences from heterogeneous behaviour. \textit{arXiv preprint arXiv:2208.09611}.

\bibitem[McFadden(1974)]{mcfadden1974conditional}
McFadden, D. (1974). Conditional logit analysis of qualitative choice behavior. In P.~Zarembka (Ed.), \textit{Frontiers in Econometrics} (pp.\ 105--142). Academic Press.

\bibitem[Train(2009)]{train2009discrete}
Train, K.E. (2009). \textit{Discrete Choice Methods with Simulation} (2nd ed.). Cambridge University Press.

\bibitem[Cao et al.(2021)]{cao2021identifiability}
Cao, Z., Tian, Y., Krishnamurthy, A., \& Schapire, R.E. (2021). Identifiability of reward functions in inverse reinforcement learning. \textit{arXiv preprint arXiv:2111.14936}.

\bibitem[Rolland et al.(2022)]{rolland2022identifiability}
Rolland, P., Viano, L., Scherrer, B., Pietquin, O., \& Cevher, V. (2022). Identifiability and generalizability from multiple experts in inverse reinforcement learning. In \textit{Advances in Neural Information Processing Systems} (pp.\ 7962--7974).

\bibitem[Wulff et al.(2023)]{wulff2023keystroke}
Wulff, D.U., Aeschbach, S., de Haan, A., \& Mata, R. (2023). A meta-analysis of keystroke dynamics for human identification and authentication. \textit{arXiv preprint arXiv:2303.04605}.

\bibitem[Ziebart et al.(2008)]{ziebart2008maximum}
Ziebart, B.D., Maas, A., Bagnell, J.A., \& Dey, A.K. (2008). Maximum entropy inverse reinforcement learning. In \textit{Proceedings of the 23rd AAAI Conference on Artificial Intelligence} (pp.\ 1433--1438).

\end{thebibliography}

\end{document}